\documentclass[fleqn,10pt]{wlscirep}
\usepackage[utf8]{inputenc}
\usepackage[T1]{fontenc}
\usepackage{booktabs}
\usepackage{tabularx}
\usepackage{xcolor}
\usepackage{tcolorbox}

\newcommand{\rev}[1]{#1}

\definecolor{mBlue}{HTML}{1F4E79}

\title{Toward an automated science of the mind}
\author[1,*]{Akshay K. Jagadish}
\author[2]{Milena Rmus}
\author[2,3]{Kristin Witte}
\author[2]{Marvin Mathony}
\author[2]{Marcel Binz}
\author[2,*]{Eric Schulz}
\affil[1]{Princeton AI Lab, Princeton University, New Jersey, USA}
\affil[2]{Institute for Human-Centered AI, Helmholtz Munich, Neuherberg, Germany}
\affil[3]{Ludwig-Maximilians-University, Munich, 80539, Germany}
\affil[*]{akshaykjagadish@gmail.com, eric.schulz@helmholtz-munich.de}

\begin{abstract}
\rev{Artificial Intelligence is starting to automate scientific discovery, raising the prospect of an automated science of the mind. We chart this field around four challenges: representing experiments, generating synthetic behavior, synthesizing models, and closing the loop to discover psychological theories.}
\end{abstract}
\begin{document}

\flushbottom
\maketitle
\thispagestyle{empty}

\section*{Introduction}
The cognitive sciences aim to understand intelligence using computational models. These models formally ground theories about cognitive processes by specifying interpretable operations and dynamics underlying human reasoning, learning, and decision-making. The standard model of discovery in the cognitive sciences typically follows a virtuous cycle of four stages \cite{klahr1988dual}. First, researchers develop a paradigm to test a particular feature of human cognition. Second, a sufficient volume of data is collected from human participants. Third, a predefined class of models is handcrafted and tested against the collected data, which corresponds to targeted hypothesis testing of the involved cognitive processes. Finally, the results are reviewed by the scientific community, who then innovate on the paradigm, model class, and model comparisons involved.

While this process has led to steady progress in our understanding of cognition, it remains constrained by structural disadvantages. The primary constraint is the inherent slowness of a cycle that relies on human involvement at every step, where the time elapsed between initial hypothesis and final publication can span years and thus decouple the pace of theory development from the velocity of data acquisition. Furthermore, this manual pipeline \rev{restricts the space of explored hypotheses to} familiar experimental designs and model classes rather than the vast landscape of possible paradigms and algorithmic solutions. This reliance on researcher-driven designs introduces subtle biases and limits the scalability of experimentation and theory development.

\begin{figure}
    \centering
    \includegraphics[width=1.\linewidth]{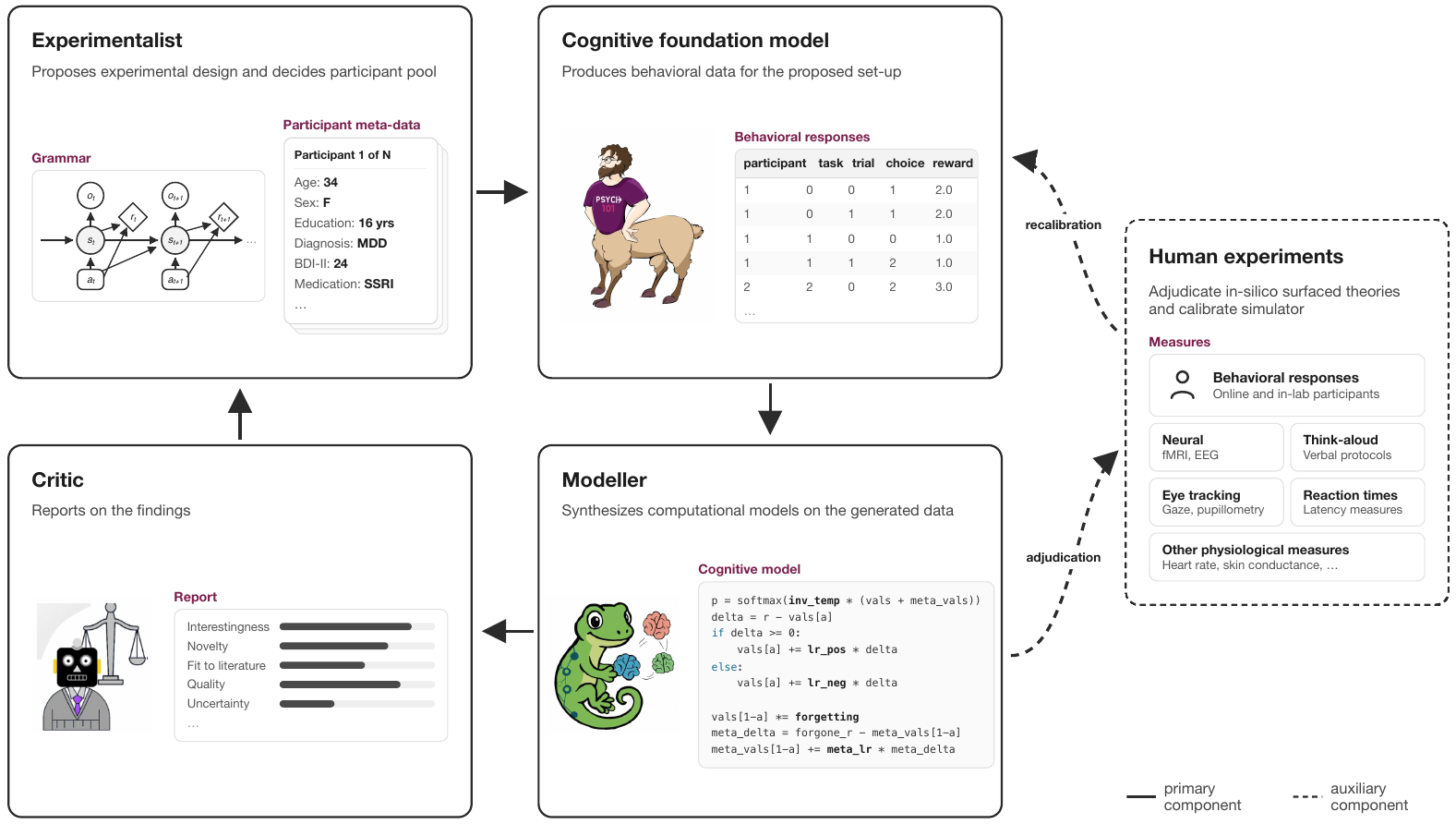}
    \caption{An automated cycle of scientific discovery in the cognitive sciences. An experimentalist proposes experiments; a cognitive foundation model simulates behavioral data; a modeller synthesizes and tests computational models; and a critic judges the ``interestingness'' of the results and passes that signal back to the experimentalist for further optimization. 
    \rev{Crucially, discoveries that survive in-silico screening are compiled into targeted human experiments---behavioral and, where relevant, verbal reports, eye-tracking, neural responses, and other physiological measurements---whose outcomes both adjudicate candidate discoveries and help recalibrate the behavioral generator, keeping the loop anchored to empirical data. Solid line indicate primary components of the cycle. Dotted line indicate auxiliary components of the cycle}}
    \label{fig:overview}
\end{figure}

These limitations raise the question: can we remove the bottlenecks imposed by human intervention and genuinely scale up scientific discovery in the cognitive sciences \cite{musslick2025automating}? We are currently at an inflection point where components capable of automating each stage of the discovery cycle are becoming available. \rev{The aim of this article is to lay out our vision for automated cognitive science.} In the following sections, we use the four stages of the cycle---\rev{representing experiments}, generating synthetic behavior, synthesizing models, and closing the loop--- \rev{as a framework for identifying opportunities and bottlenecks in scaling up scientific discovery in cognitive science (Figure~\ref{fig:overview} and the Box~1)}.
\rev{We then consider the epistemic challenges that arise when the search for cognitive theories itself becomes increasingly automated} and finally, conclude with how automated science of the mind is no longer a distant fiction but an imminent reality.

\section*{\rev{Representing experiments}}

Automated discovery begins by defining the space of possible experiments. If machine algorithms are to generate new paradigms rather than merely optimize existing ones, a formal language for describing experimental structures and how their constituent components can be combined is required. A natural starting point is generative grammars over experiments. For instance, a grammar over Markov Decision Processes (MDPs) could provide an initial foundation. Because MDPs are formally defined by environmental states, transition dynamics, and reward structures, they can capture a vast class of decision-making tasks used in psychology and neuroscience, ranging from simple multi-armed bandits to multi-step planning paradigms.

Yet, this approach immediately exposes a representational bottleneck: automated discovery is fundamentally constrained by the expressivity of its underlying grammar. \rev{Large parts of cognitive science are not naturally expressed as MDPs: psychophysics and perception require continuous, renderable stimulus spaces; memory paradigms employ study--test structures; language comprehension involves open-ended linguistic stimuli; and social cognition operates through multi-agent interactions. 
Each family suggests a different representational strategy: domain-specific grammars, experiments expressed as executable programs, or natural language from which machine-readable specifications are compiled. The central challenge for this field is therefore not to construct a single universal grammar of cognition, but to develop representations expressive enough to span diverse cognitive phenomena while remaining structured enough to support systematic search.
} 

Instead of committing to a representation in advance, we propose using an LLM itself as an intelligent experiment sampler. In this framework, an LLM directly proposes a set of experiments that are best positioned to elicit a specific type of behavior or test a specific hypothesis, and critically, it can further refine them based on feedback, adding or removing key experimental factors as needed. \rev{Such samplers extend a longer tradition of automated experimental design, from adaptive optimal-design methods to closed-loop platforms that integrate design, data collection, and analysis \cite{musslick2025automating}.}

\section*{Generating \rev{synthetic behavior}}

Once a task is specified, the cycle proceeds to the generation of synthetic behavioral data. \rev{Several approaches could supply the behavioral substrate for this stage, including task-specific simulators, ensembles of fitted cognitive models, generative agents, and foundation models trained on large corpora of human behavior. Recent iterations of the \textit{Centaur} model \cite{binz2025foundation}---a foundation model of human cognition---provide one example of the latter, generating data for any cognitive experiment that can be expressed in language format.} Unlike traditional cognitive models that require manual parameter tuning, foundation models, like Centaur, can generate \textit{de novo} data that are \rev{often difficult to distinguish} from human behavior. 
While the ability of these models to generalize to entirely novel experimental structures remains a subject of ongoing debates, current evidence suggests they can (at least to some extent) successfully extrapolate to tasks outside their initial training distribution.

\rev{In an ongoing work, we are expanding the behavioral dataset used for fine-tuning to capture additional cognitive domains and diverse experimental structures} and are also integrating extensive subject-specific metadata into the fine-tuning process. By incorporating demographics (e.g., age, gender, nationality) and questionnaire scores (e.g., psychiatric symptom scales), the data generation process can be conditioned on specific individual differences. This allows the system to simulate not just a generic agent, but a targeted participant profile, for instance, simulating the behavior of a 30-year-old individual with high scores in obsessive-compulsive traits performing a specific planning task. 
By parsing both the proposed experiment and these subject-specific features into the generative engine, we can produce high-fidelity behavioral data that reflect the high-dimensional complexity of real-world populations. 
\rev{Going beyond behavior, there are foundation models that can predict human brain activity across a variety of naturalistic and experimental conditions \cite{d2026foundation}, pointing toward a future where generators can simulate behavior and its neural signatures. In the end, whichever simulator is used, they must be continually calibrated against held-out human data and upon careful vetting, they} are submitted to the next step of the cycle: model synthesis.

\section*{Synthesizing models}

The third stage, model synthesis and comparison, is traditionally labor-intensive: researchers manually specify a small set of candidate models, derive their likelihood functions, and fit them to data using computationally demanding optimization routines. The process is slow and inherently biased toward the narrow set of ``cognitively plausible'' functions that humans are familiar with and capable of formalizing. Automating this stage means shifting from carefully \textit{handcrafting} to \textit{synthesizing} models at scale, over a search space that encompasses a vast landscape of algorithmic hypotheses.
 
\rev{Several families of approaches to this problem are emerging \cite{rmusgenerating, castro2025discovering, weinhardt2025automated}. 
For instance, LLM-guided program synthesis methods such as} Guided generation of Computational Cognitive Models (GeCCo) \cite{rmusgenerating} can synthesize cognitive models as Python functions and iteratively refine them based on feedback regarding their predictive performance while maintaining a record of all intermediate hypotheses. 
\rev{In contrast, Sparse and Interpretable Cognitive Equations (SPICE) \cite{weinhardt2025automated} uses a combination of task-specific neural networks that are trained to predict human behavior and a sparse equation discovery algorithm to discover interpretable models of cognitive mechanisms. 
While LLM-based search is very general, it inherits biases from its training data; symbolic regression methods are data-driven but restricted in expressivity and generality. A potential hybrid of the two offers a path toward discovering mechanisms that human intuition might overlook, transforming model synthesis into an objective, high-throughput search over cognitive programs.}

\rev{Irrespective of the approach we take, a key challenge these methods face is identifiability: automated search can generate distinct programs that are behaviorally indistinguishable. One way to mitigate this issue is to impose additional constraints during the search process. For example, neural data can offer a natural adjudicator: synthesized models make explicit commitments about latent variables, such as subjective values or prediction errors, that can be tested against neural recordings. Requiring candidate models to predict both choices and neural signatures can not only help distinguish among behaviorally equivalent mechanisms but also potentially bridge Marr's algorithmic level with implementation.}

\section*{Closing the loop}

The final---and conceptually hardest---step is to close the loop: deciding how the outcome of one iteration should shape the next experiment, participant profiles, and model search. In a fully autonomous pipeline, this requires specifying an objective for the process of scientific discovery itself. Classical strategies from optimal experimental design and active learning may offer a principled starting point: we can propose tasks that maximize expected information gain, reduce posterior uncertainty over models, or target regions of the task grammar where candidate explanations disagree most strongly. Yet these criteria are myopic. They reward \emph{discriminability} even when the resulting distinctions are scientifically unilluminating, and they often steer the system toward edge cases that are statistically informative but theoretically dull.
 
To move beyond this, the loop must optimize not only informativeness, but also \emph{interestingness}. Building on recent work on open-ended discovery and self-improving search \cite{zhangomni}, we can introduce a critic model that evaluates the outcome of each cycle and proposes what to do at the beginning of the next cycle, following an LLM-as-judge framework. In concrete terms, the critic scores a \emph{discovery tuple}---the proposed experiment, the simulated participant profile(s), the generated data, and the space of models that best explains it---along dimensions such as novelty relative to the system's prior discoveries, compressibility or simplicity of the best explanation, the presence of qualitative signatures, and the extent to which the results suggest a transferable principle rather than a task-specific hack.
 
Closing the loop, experiment generation is guided by an \emph{interestingness signal} that rewards discoveries with high conceptual yield. The system can then iteratively bias its proposals toward configurations that produce sharp theoretical contrasts, uncover new behavioral phenomena, or force the invention of new explanatory primitives. Importantly, this does not eliminate the scientist. Instead, it acts as a high-throughput search in the hypothesis space: humans define the problem, the expressiveness of the task language, the structure of the output model family, evaluation principles, the shape of the expected theory, and other constraints, while the automated loop surfaces a small set of unusually informative experiments and mechanisms. The highest-scoring discoveries can then be validated \textit{in vivo} with human participants\rev{---behaviorally and with neural recording---}keeping the process anchored to biological cognition while dramatically scaling up and accelerating the search for theory. \rev{The resulting data serve two functions: they adjudicate the candidate discovery, and they stream back into the training and calibration corpus of the behavioral generator, so that the in-silico engine tracks real behavior rather than drifting from it. Empirical grounding is thus not the end point; it kicks off the calibration phase that precedes the start of every new cycle (Figure~\ref{fig:overview}).}

 \begin{tcolorbox}[sharp corners, colback=mBlue!5!white,colframe=mBlue!75!black, width=1.\textwidth, left=4pt,right=4pt, top=4pt, bottom=4pt, title=\textbf{Surveying the landscape of automated cognitive science.} \label{box:landscape}]
 
Each stage of the discovery cycle admits multiple instantiations, and each comes with its own open challenges. 

\begin{enumerate}
    \item \textbf{Representing experiments:} we proposed using generative grammars over task families (e.g., MDPs) with LLMs as experiment samplers; alternatives include adaptive optimal experimental design, closed-loop design--collection--analysis platforms \cite{musslick2025automating}, and experiments expressed as executable programs. The key challenges here are representing non-MDP domains (perception, memory, language, social cognition, emotion), trading off coverage against ease of search, and automated validity checking. 
    
    \item \textbf{Generating synthetic behavior : }we build on cognitive foundation models such as Centaur \cite{binz2025foundation}, while task-specific simulators and ensembles of fitted cognitive models offer alternatives; open challenges include heuristic shortcuts \cite{xie2025centaur}, bias towards the WEIRD population \cite{norenzayan2010weirdest}, continual calibration against human benchmarks, and joint behavioral--neural response generation \cite{d2026foundation}.
    
    \item \textbf{Synthesizing models:} we rely on LLM-guided program synthesis (e.g., GeCCo \cite{rmusgenerating}), with evolutionary search over symbolic programs \cite{castro2025discovering} and Sparse and Interpretable Cognitive Equations \cite{weinhardt2025automated} as alternatives; the challenges are biases towards LLM priors, model sensitivity, a rugged optimization space, behavioral identifiability, jointly predicting neural and behavioral responses, and metric gaming. 

    \item \textbf{Closing the loop:} we propose critic-guided interestingness (LLM-as-judge) \cite{zhangomni}, with information-gain and active-learning objectives as alternatives; defining and measuring scientific value, critic gaming, multi-objective search, and norms for human oversight remain open.
\end{enumerate}

\end{tcolorbox}


\section*{Discussion}
Jorge Luis Borges imagined the \textit{Library of Babel}: an infinite archive containing every possible book, masterpieces and proofs alongside endless gibberish. An automated science of the mind risks building the cognitive analogue of that library: a vast enumerative engine that generates every possible task, every possible dataset, every possible model, most of it worthless. The goal here, however, is not to flood cognitive science with an ocean of synthetic ``books''. It is to surface the rare volumes that matter: experiments that carve nature at its joints, models that compress behavior into explanatory principles, and predictions that survive contact with actual data. If automation succeeds, it should not remove the scientist from discovery, but rather act as an amplifier of our ability to search, to falsify, and to learn.
 
This paper has outlined \rev{one possible} end-to-end vision for such a system: an engine that proposes experiments, generates behavioral data in silico, discovers mechanistic models at scale, and closes the loop by steering search toward high-value discoveries \rev{(Box~1)}. The promise is apparent, and so are the failure modes. A credible discussion must therefore be blunt about what can go wrong---because the central risk is not merely technical failure, but epistemic failure: producing persuasive-looking results that are scientifically hollow.
 
The proposed cycle is only as strong as its weakest step.
First, automated experimental design is bounded by the expressibility and interpretability of the task language. A task grammar determines what experiments can be proposed, and much of scientific innovation comes from stepping outside the currently explored space: inventing new paradigms, new measurements, or new representational framings. No search procedure can discover what the representation forbids. This makes the experiment grammar itself a central scientific object: it must be able to \rev{span diverse cognitive task domains without being constrained to any impoverished formalisms, and it should be probed to find its failure modes and systematically expanded.}
 
Second, synthetic data generation is not guaranteed to be a faithful substitute for humans. Behavioral foundation models may rely on shortcuts, \rev{exhibit bias towards behavior from the WEIRD\footnote{Western, Educated, Industrialized, Rich, and Democratic \cite{norenzayan2010weirdest}} population}, drift toward ``average'' subjects, or fail under genuine distribution shifts. Even plausible-looking behavior can be produced for the wrong reasons, yielding an illusion of mechanistic clarity.
A pragmatic stance is to treat such models as accelerators for hypothesis generation rather than as ground truth: they should be continuously calibrated against benchmark effects, challenged with adversarial tasks designed to expose heuristics, and, where appropriate, complemented or replaced by smaller, local simulators tuned to specific task families.

Third, model discovery over programs is not smooth optimization. Instead, the optimization space is rugged: tiny edits to the code can break a model, distinct mechanisms can be behaviorally indistinguishable, and LLM-guided search can inherit strong priors and stylistic biases from its training data. There is no guarantee of finding a globally best mechanism, and evaluation metrics can be gamed unless carefully designed. Nevertheless, it is undeniable that this approach can explore
far more candidates than was previously possible, and early results suggest it can discover new mechanistic descriptions rather than rephrasing familiar ones.

Finally, closing the loop with an ``interestingness'' signal is both enticing and dangerous. A critic could reward theatrical novelty, drift toward idiosyncratic edge cases, or be gamed by policies that learn to manufacture superficial weirdness. Interestingness should therefore not be a single scalar oracle. A safer design is multi-objective: combine novelty with robustness, parsimony, generalization across tasks and populations, and, crucially, unification, i.e., explaining multiple phenomena with shared mechanisms. \rev{How to quantify and weigh these objectives, and when a result constitutes genuine progress, are open questions that demand substantial focus from future research.}

\rev{Overall, these risks point toward the same conclusion: automation does not remove the scientist}, it shifts human effort from enumerating possibilities to defining problem representations, the shape of the solutions, constraints, and decisive tests.
\rev{Specifically, we see at least three indispensable roles for researchers. 
First, \emph{research taste}: posing challenging questions and specifying the shape of expected solution. 
Second, \emph{representation}: designing domain-specific languages for models, experiments, and evaluation criteria that keep automated search scientifically meaningful, and curating what counts as progress. 
Third, \emph{grounding}: collecting rich human data from naturalistic, longitudinal experiments with dense measurements -- ranging from verbal reports to neural recordings -- that allow continual calibration of behavioral generators, constraint the space of possible theories, and provide a testing ground for ultimately verifying surfaced theories.} 
In this view, machines search and compare possibilities at scales humans cannot, while humans define what is worth asking and what counts as an explanation.

\rev{If these challenges are met, the upside is large.} Automating the loop could \rev{substantially increase the throughput of discovery while providing a full record of its search trajectory that make its conclusions auditable}. \rev{It promises mechanistic accounts of individual differences rather than the mean population, tighter integration of behavioral theory with neural data, and a route to addressing the generalization crisis: by explicitly searching for invariances across task variants and populations, generalization becomes a selection criterion for theories rather than a post-hoc hope.}

\rev{Overall, we appear to be near an inflection point where the components of an automated science of the mind are simultaneously becoming feasible.} The challenge now is epistemic as much as technical: to build systems that produce fewer books and more knowledge, and to develop norms that keep automated discovery tethered to empirical validation. The Library of Babel contains everything; science is the art of finding the few pages worth reading. An automated discovery engine will be worth building if it reliably elevates those pages.

\section*{Acknowledgements}
AJ is supported by the Natural and Artificial Mind (NAM) Fellowship, which is generously supported by the Scully Peretsman Foundation.
ES, MR, KW, MM, and MB are funded by the Helmholtz Association, an ERC Starting Grant on "Towards an Artificial Cognitive Science'', and a Wellcome Discovery Award.

\bibliography{sample}

\end{document}